\documentclass[conference]{IEEEtran}

\usepackage{graphicx}
\usepackage{tabularx}
\usepackage{amssymb}
\usepackage{booktabs} 
\usepackage{epstopdf} 
\usepackage{subfigure} 
\usepackage{multirow}
\usepackage{amsthm}
\usepackage{lipsum}
\usepackage{stmaryrd}
\usepackage{makecell}
\usepackage{multirow}
\usepackage{hhline}
\usepackage{cellspace}
\usepackage{amsmath}
\usepackage{hyperref}

\usepackage{comment}

\ifCLASSINFOpdf
\else
\fi
\newcolumntype{M}[1]{>{\centering\arraybackslash}m{#1}}

\begin{document}
%

\title{TI-CNN: Convolutional Neural Networks for Fake News Detection}

\author{\IEEEauthorblockN{Yang Yang}
 \IEEEauthorblockA{Beihang University\\
Beijing, China\\
 Email: yangyangfuture@gmail.com}
 \and
 \IEEEauthorblockN{Lei Zheng}
 \IEEEauthorblockA{University of Illinois at Chicago\\
 Chicago, United States\\
 Email: lzheng21@uic.edu}
 \and
 \IEEEauthorblockN{Jiawei Zhang}
 \IEEEauthorblockA{Florida State University\\
 Florida, United States\\
 Email: jzhang@cs.fsu.edu}
 \and
 \IEEEauthorblockN{Qingcai Cui}
 \IEEEauthorblockA{Beihang University\\
	Beijing, China\\
 Email: cqc@cuiqingcai.com}
  \and
 \IEEEauthorblockN{Xiaoming Zhang}
 \IEEEauthorblockA{Beihang University\\
	Beijing, China\\
 Email: yolixs@163.com}
  \and
 \IEEEauthorblockN{Zhoujun Li}
 \IEEEauthorblockA{Beihang University\\
	Beijing, China\\
 Email: lizj@buaa.edu.cn}
   \and
 \IEEEauthorblockN{Philip S. Yu}
 \IEEEauthorblockA{University of Illinois at Chicago\\
 Chicago, United States\\
 Email: psyu@uic.edu}
 }
\maketitle

\begin{abstract}
With the development of social networks, fake news for various commercial and political purposes has been appearing in large numbers and gotten widespread in the online world. With deceptive words, people can get infected by the fake news very easily and will share them without any fact-checking. For instance, during the 2016 US president election, various kinds of fake news about the candidates widely spread through both official news media and the online social networks. These fake news is usually released to either smear the opponents or support the candidate on their side. The erroneous information in the fake news is usually written to motivate the voters' irrational emotion and enthusiasm. Such kinds of fake news sometimes can bring about devastating effects, and an important goal in improving the credibility of online social networks is to identify the fake news timely. In this paper, we propose to study the ``fake news detection'' problem. Automatic fake news identification is extremely hard, since pure model based fact-checking for news is still an open problem, and few existing models can be applied to solve the problem. With a thorough investigation of a fake news data, lots of useful explicit features are identified from both the text words and images used in the fake news. Besides the explicit features, there also exist some hidden patterns in the words and images used in fake news, which can be captured with a set of latent features extracted via the multiple convolutional layers in our model. A model named as TI-CNN (\underline{T}ext and \underline{I}mage information based \underline{C}onvolutinal \underline{N}eural \underline{N}etwork) is proposed in this paper. By projecting the explicit and latent features into a unified feature space, TI-CNN is trained with both the text and image information simultaneously. Extensive experiments carried on the real-world fake news datasets have demonstrate the effectiveness of TI-CNN in solving the fake new detection problem.
\end{abstract}


%
\IEEEpeerreviewmaketitle

\section{Introduction}
Fake news is written in an intentional and unverifiable language to mislead readers. It has a long history since the 19th century. In 1835, New York Sun published a series of articles about ``the discovery of life on the moon". Soon the fake stories were printed in newspapers in Europe. Similarly, fake news widely exists in our daily life and is becoming more widespread following the Internet's development. Exposed to the fast-food culture, people nowadays can easily believe something without even checking whether the information is correct or not, such as the ``FBI agent suspected in Hillary email leaks found dead in apparent murder-suicide''. These fake news frequently appear during the United States presidential election campaign in 2016. This phenomenon has aroused the attention of people, and it has a significant impact on the election. 

Fake news dissemination is very common in social networks \cite{allcott2017social}. Due to the extensive social connections among users, fake news on certain topics, e.g., politics, celebrities and product promotions, can propagate and lead to a large number of nodes reporting the same (incorrect) observations rapidly in online social networks. According to the statistical results reported by the researchers in Stanford University, 72.3\% of the fake news actually originates from the official news media and online social networks \cite{conroy2015automatic}. The potential reasons are provided as follows. Firstly, the emergence of social media greatly lower down the barriers to enter in the media industry. Various online blogs, ``we media", and virtual communities are becoming more and more popular in recent years, in which everyone can post news articles online. Secondly, the large number of social media users provide a breeding ground for fake news. Fake news involving conspiracy and pitfalls can always attract our attention. People like to share this kind of information to their friends. Thirdly, the `trust and confidence' in the mass media greatly dropped these years. More and more people tend to trust the fake news by browsing the headlines only without reading the content at all.


Fake news identification from online social media is extremely challenging due to various reasons. Firstly, it's difficult to collect the fake news data, and it is also hard to label fake news manually \cite{wang2017liar}. News that appears on Facebook and Twitter news feeds belongs to private data. To this context so far, few large-scale fake news detection public dataset really exists. Some news datasets available online involve a small number of the instances only, which are not sufficient to train a generalized model for application. Secondly, fake news is written by human. Most liars tend to use their language strategically to avoid being caught. In spite of the attempt to control what they are saying, language “leakage” occurs with certain verbal aspects that are hard to monitor such as frequencies and patterns of pronoun, conjunction, and negative emotion word usage \cite{feng2013detecting}. Thirdly, the limited data representation of texts is a bottleneck of fake news identification. In the bag-of-words approach, individual words or ``n-grams'' (multiword) frequencies are aggregated and analyzed to reveal cues of deception. Further tagging of words into respective lexical cues for example, parts of speech or ``shallow syntax'' \cite{markowitz2016linguistic}, affective dimensions \cite{vrij2006empirical}, and location-based words \cite{ott2013negative} can all provide frequency sets to reveal linguistic cues of deception \cite{newman2003lying,hancock2007lying}. The simplicity of this representation also leads to its biggest shortcoming. In addition to relying exclusively on language, the method relies on isolated n-grams, often divorced from useful context information. Word embedding techniques provide a useful way to represent the meaning of the word. In some circumstances, sentences of different lengths can be represented as a tensor with different dimensions. Traditional models cannot handle the sparse and high order features very well. 

\begin{figure}[ht]
\centering
\subfigure[b][Cartoon in fake news.]{\label{fig:cartoon}
\includegraphics[width=0.22\textwidth,height=2.5cm]{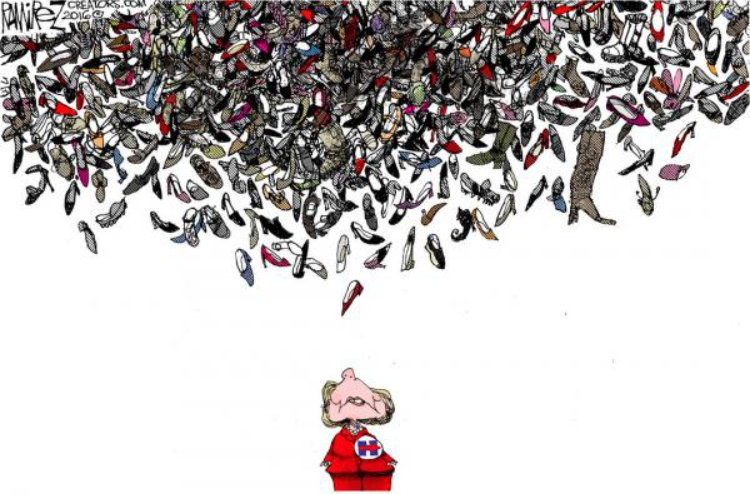}}
\subfigure[b][Altered low-resolution image.]{\label{fig:obama}
\includegraphics[width=0.22\textwidth,height=2.5cm]{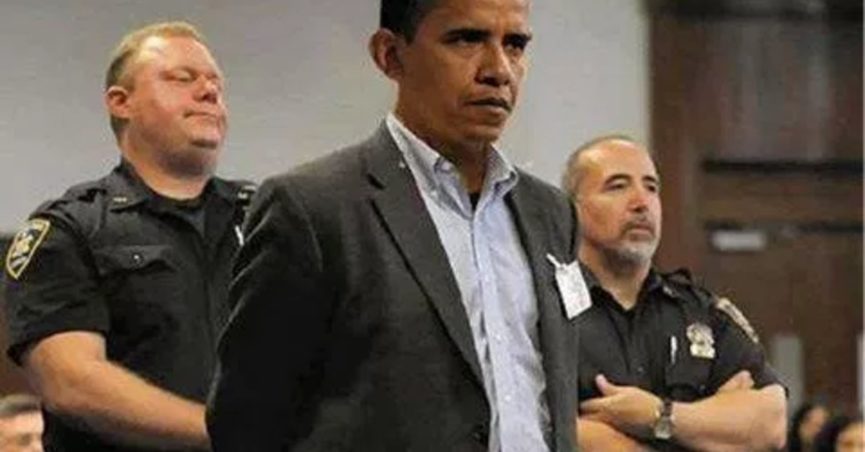}}
\subfigure[b][Irrelevant image in fake news.]{\label{fig:amish}
\includegraphics[width=0.22\textwidth,height=2.5cm]{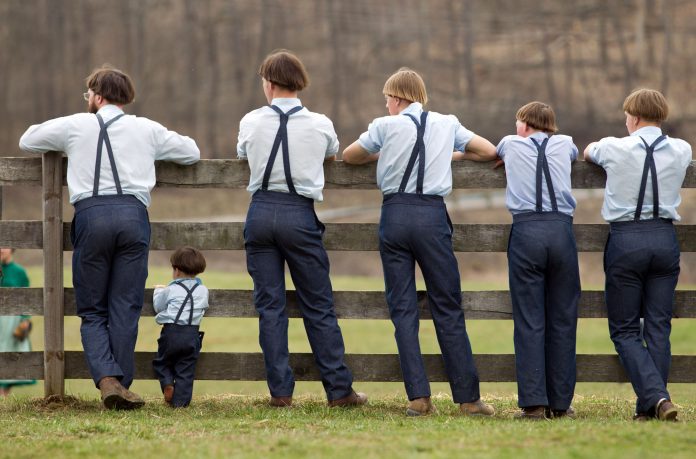}}
\subfigure[b][Low-resolution image.]{\label{fig:hillary}
\includegraphics[width=0.22\textwidth,height=2.5cm]{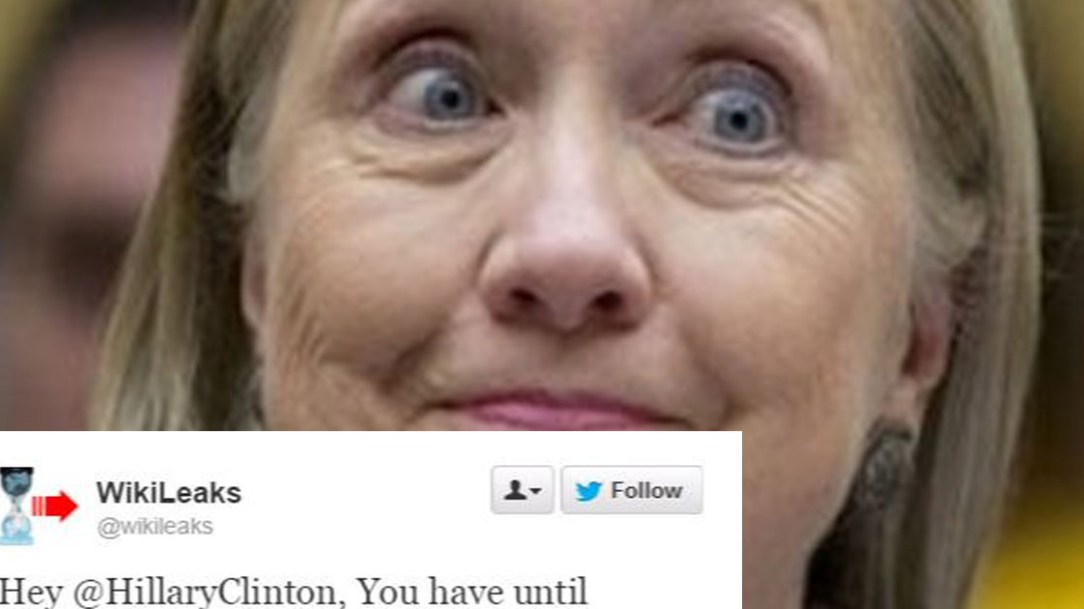}}
\label{fig:fake_image}
\caption{The images in fake news: (a) `FBI Finds Previously Unseen Hillary Clinton Emails On Weiner's Laptop', (b)`BREAKING: Leaked Picture Of Obama Being Dragged Before A Judge In Handcuffs For Wiretapping Trump', (c) `The Amish Brotherhood have endorsed Donald Trump for president', (d) `Wikileaks Gives Hillary An Ultimatum: QUIT, Or We Dump Something Life-Destroying'. The news texts of images (c) and (d) are represented in Section \ref{ssec:case_study}}
\end{figure}

Though the deceivers make great efforts in polishing fake news to avoid being found, there are some leakages according to our analysis from the text and image aspect respectively. For instance, the lexical diversity and cognition of the deceivers are totally different from the truth teller. Beyond the text information, images in fake news are also different from that in real news. As shown in Fig. \ref{fig:fake_image}, cartoons, irrelevant images (mismatch of text and image, no face in political news) and altered low-resolution images are frequently observed in fake news. In this paper, we propose a TI-CNN model to consider both text and image information in fake news detection. Beyond the explicit features extracted from the data, as the development of the representative learning, convolutional neural networks are employed to learn the latent features which cannot be captured by the explicit features. Finally, we utilize TI-CNN to combine the explicit and latent features of text and image information into a unified feature space, and then use the learned features to identify the fake news. Hence, the contributions of this paper are summarized as follows:
\begin{itemize}
\item We collect a high quality dataset and take in-depth analysis on the text from multiple perspectives.
\item Image information is proved to be effective features in identifying the fake news.
\item A unified model is proposed to analyze the text and image information using the covolutoinal neural networks.
\item The model proposed in this paper is an effective way to recognize fake news from lots of online information.
\end{itemize}

In the rest of the paper, we first define the problem of fake news identification. Then we introduce the analysis on the fake news data. A unified model is proposed to illustrate how to model the explicit and latent features of text and image information. The details of experiment setup is demonstrated in the experiment part. At last, we compare our model with several popular methods to show the effectiveness of our model.

\section{Related Work}
Deception detection is a hot topic in the past few years. Deception information includes scientific fraud, fake news, false tweets etc.
Fake news detection is a subtopic in this area. Researchers solve the deception detection problem from two aspects: 1) linguistic approach. 2) network approach. 
\subsection{Linguistic approaches}
Mihalcea and Strapparvva 2009 \cite{mihalcea2009lie} started to use natural language processing techniques to solve this problem. Bing Liu et.al. \cite{hu2004mining} analyzed fake reviews on Amazon these years based on the sentiment analysis, lexical, content similarity, style similarity and semantic inconsistency to identify the fake reviews. Hai et al. \cite{hai2016deceptive} proposed semi-supervised learning method to detect deceptive text on crowdsourced datasets in 2016. 

The methods based on word analysis is not enough to identify deception. Many researchers focus on some deeper language structures, such as the syntax tree. In this case, the sentences are represented as a parse tree to describe syntax structure, for example noun and verb phrases, which are in turn rewritten by their syntactic constituent parts \cite{feng2012syntactic}.


\subsection{Network-based approaches}
Another way to identify the deception is to analyze the network structure and behaviors, which are important complementary features. As the development of knowledge graph, it will be very helpful to check fact based on the relationship among entities. Ciampaglia et al. \cite{ciampaglia2015computational} proposed a new concept `network effect' variables to derive the probabilities of news. The methods based on the knowledge graph analysis can achieve 61\% to 95\% accuracy. Another promising research direction is exploiting the social network behavior to identify the deception. 

\subsection{Neural Network based approaches}
Deep learning models are widely used in both academic community and industry. In computer vision \cite{krizhevsky2012imagenet} and speech recognition \cite{graves2013speech}, the state-of-art methods are almost all deep neural networks. In the natural language processing (NLP) area, deep learning models are used to train a model that can represent words as vectors. Then researchers propose many deep learning models based on the word vectors for QA \cite{chen2015abc} and summarization\cite{kaikhah2004automatic}, etc. Convolutional neural networks (CNN) utilize filters to capture the local structures of the image, which performs very well on computer vision tasks. Researchers also find that CNN is effective on many NLP tasks. For instance, semantic parsing \cite{yih2014semantic}, sentence modeling \cite{kalchbrenner2014convolutional}, and other traditional NLP tasks \cite{collobert2011natural}.

\begin{figure}[ht]
\begin{center}
\includegraphics[width=0.5\textwidth]{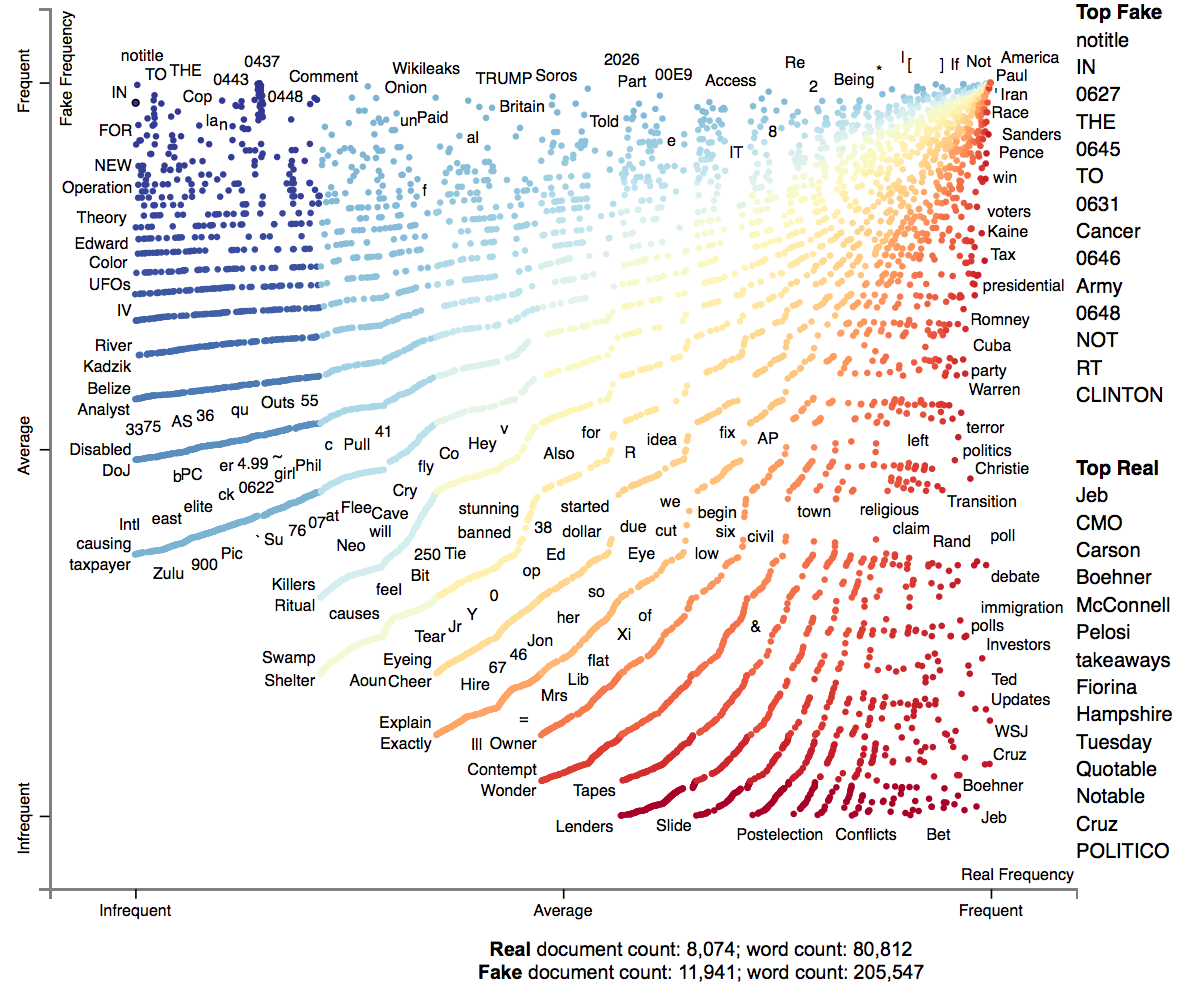}
\end{center}
\caption{Word frequency in titles of real and fake news. If the news has no title, we set the title as `notitle'. The words on the top-left are frequently used in fake news, while the words on the bottom-right are frequently used in real news.  The `Top Fake' words are capital characters and some meaningless numbers that represent special characters, while the `Top Real' words contain many names and motion verbs, i.e., `who' and `what' --- the two important factors in the five elements of news: when, where, what, why and who.}
\label{fig:word_frequency}
\end{figure}

\section{Problem Definition}


Given a set of $m$ news articles containing the text and image information, we can represent the data as a set of text-image tuples $\mathcal{A} = \{(A_i^T, A_i^I)\}_i^m$. In the fake news detection problem, we want to predict whether the news articles in $\mathcal{A}$ are fake news or not. We can represent the label set as $\mathcal{Y} = \{[1,0], [0,1]\}$, where $[1,0]$ denotes real news while $[0,1]$ represents the fake news. Meanwhile, based on the news articles, e.g., $(A_i^T, A_i^I) \in \mathcal{A}$, a set of features (including both explicit and latent features to be introduced later in Model Section) can be extracted from both the text and image information available in the article, which can be represented as $\mathbf{X}_i^T$ and $\mathbf{X}_i^I$ respectively. The objective of the \textit{fake news detection} problem is to build a model $f: \{\mathbf{X}_i^T, \mathbf{X}_i^I\}_i^m \in \mathbb{X} \to \mathcal{Y}$ to infer the potential labels of the news articles in $\mathcal{A}$.


\section{Data Analysis} 
To examine the finding from the raw data, a thorough investigation has been carried out to study the text and image information in news articles. There are some differences between real and fake news on American presidential election in 2016. We investigate the text and image information from various perspectives, such as the computational linguistic, sentiment analysis, psychological analysis and other image related features. We show the quantitative information of the data in this section, which are important clues for us to identify fake news from a large amount of data.

\subsection{Dataset}
The dataset in this paper contains 20,015 news, i.e., 11,941 fake news and 8,074 real news. It is available on google drive\footnote{\href{https://drive.google.com/file/d/0B3e3qZpPtccsMFo5bk9Ib3VCc2c/view?usp=sharing&resourcekey=0-_eqAfKOCKbuE-xFFCmEzyg}{google drive link}}. If it is not available, you can also get the data set from one drive \footnote{\href{https://1drv.ms/u/s!ArqIn8rhvqAbgT5hQR9PakJWxh53?e=7xADFd}{one drive link}}. For fake news, it contains text and metadata scraped from more than 240 websites by the Megan Risdal on Kaggle\footnote{https://www.kaggle.com/mrisdal/fake-news}. The real news is crawled from the well known authoritative news websites, i.e., the New York Times, Washington Post, etc. The dataset contains multiple information, such as the title, text, image, author and website. To reveal the intrinsic differences between real and fake news, we solely use the title, text and image information.

\begin{figure*}[ht]
\centering
\subfigure[b][The number of words in news.]{\label{fig:words}
\includegraphics[width=0.23\textwidth]{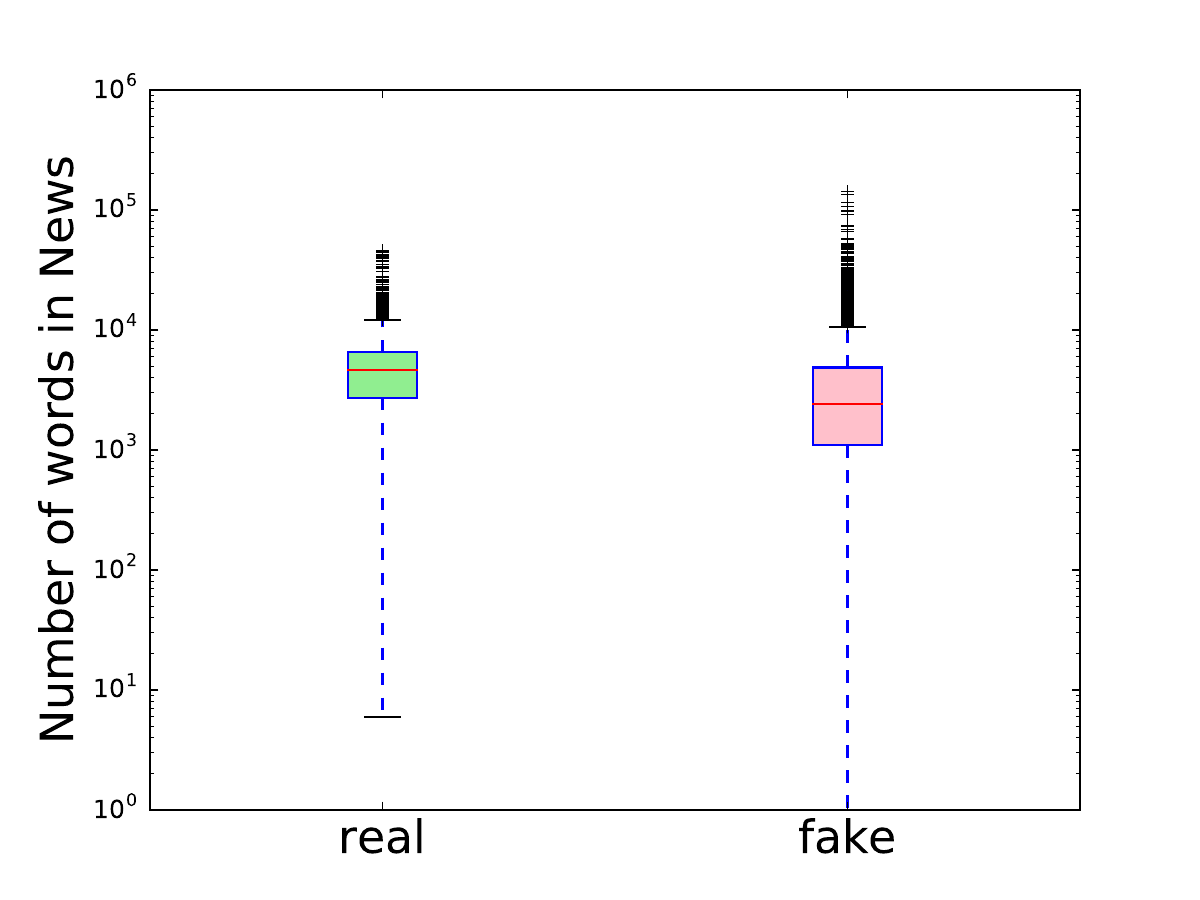}}
\subfigure[b][The average number of words in a sentence.]{\label{fig:avg_words}\includegraphics[width=0.23\textwidth]{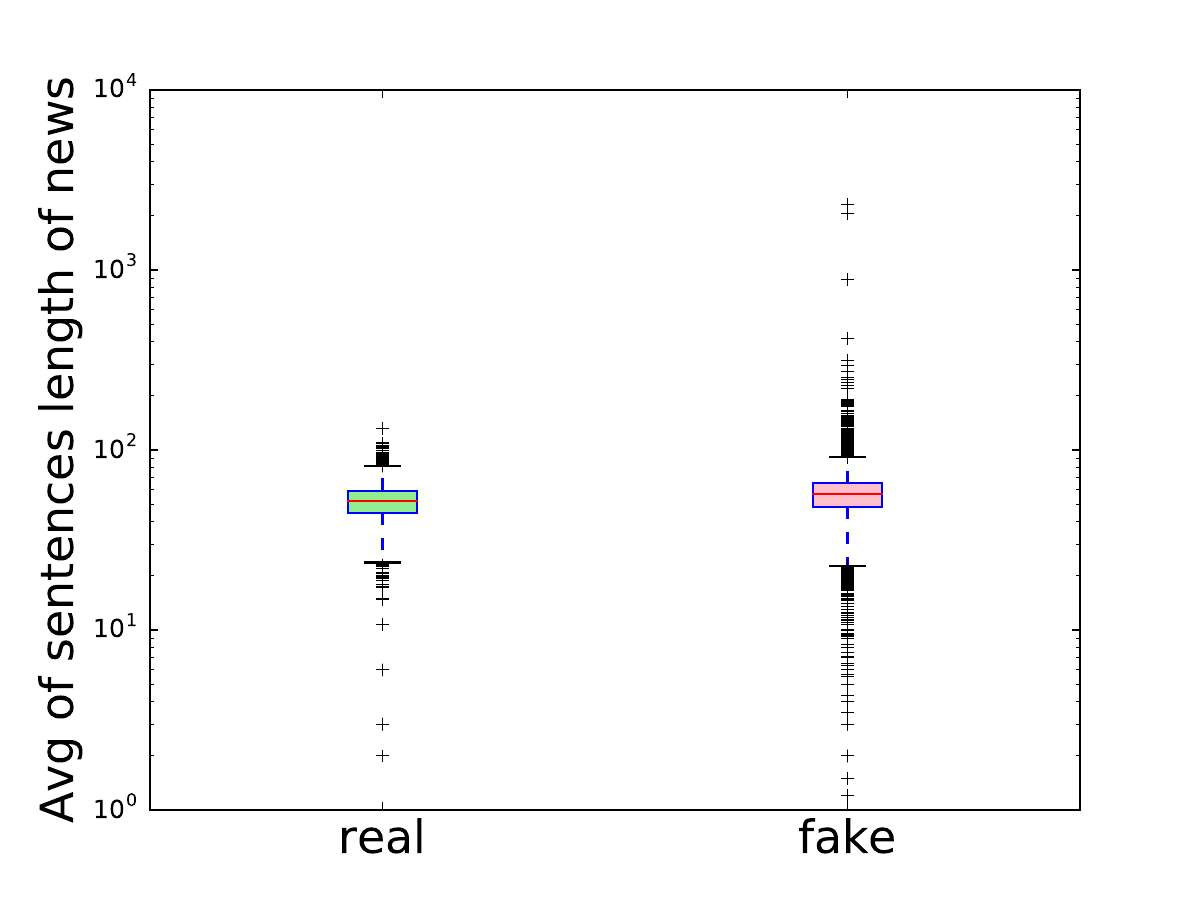}}
\subfigure[b][Question mark in news.]{\label{fig:question_mark}
\includegraphics[width=0.23\textwidth]{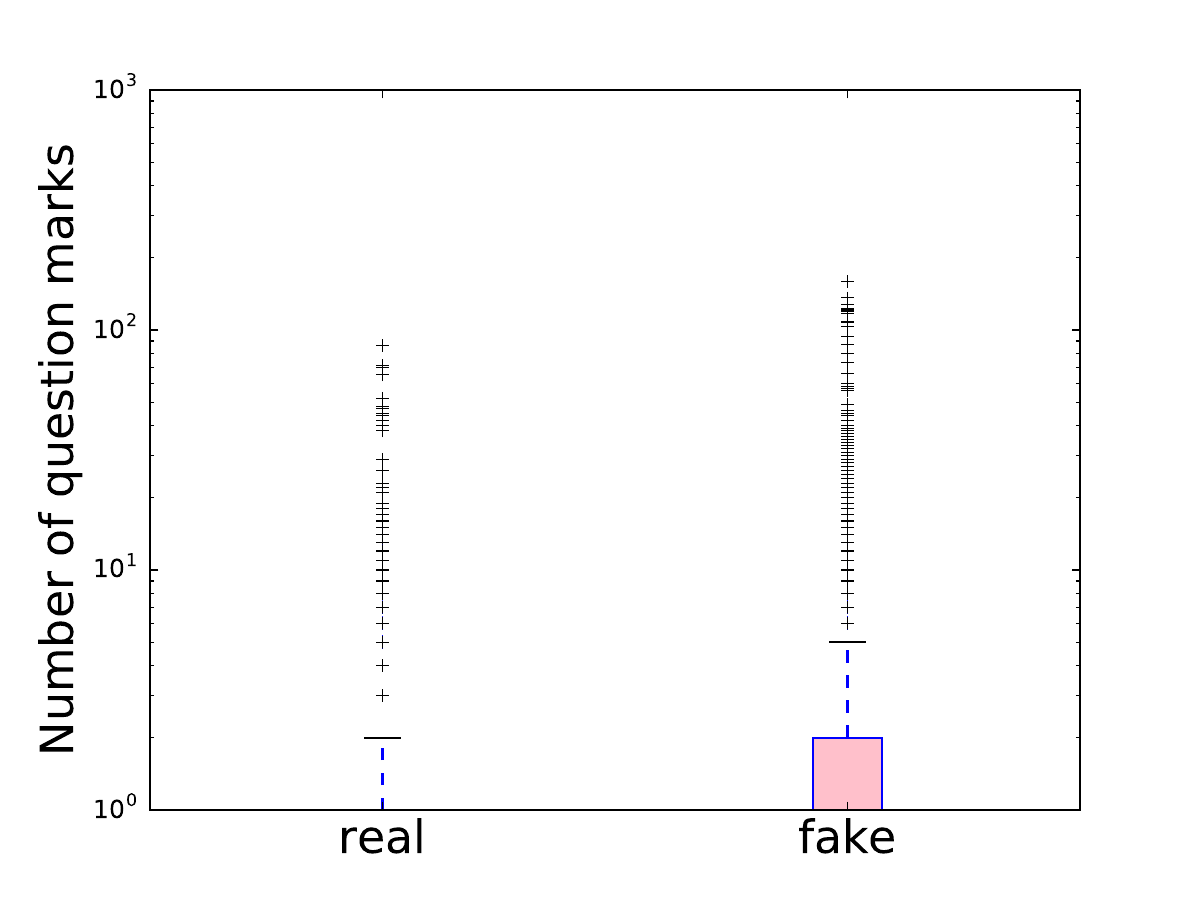}}
\subfigure[b][Exclamation mark in news.]{\label{fig:exclamation}
\includegraphics[width=0.23\textwidth]{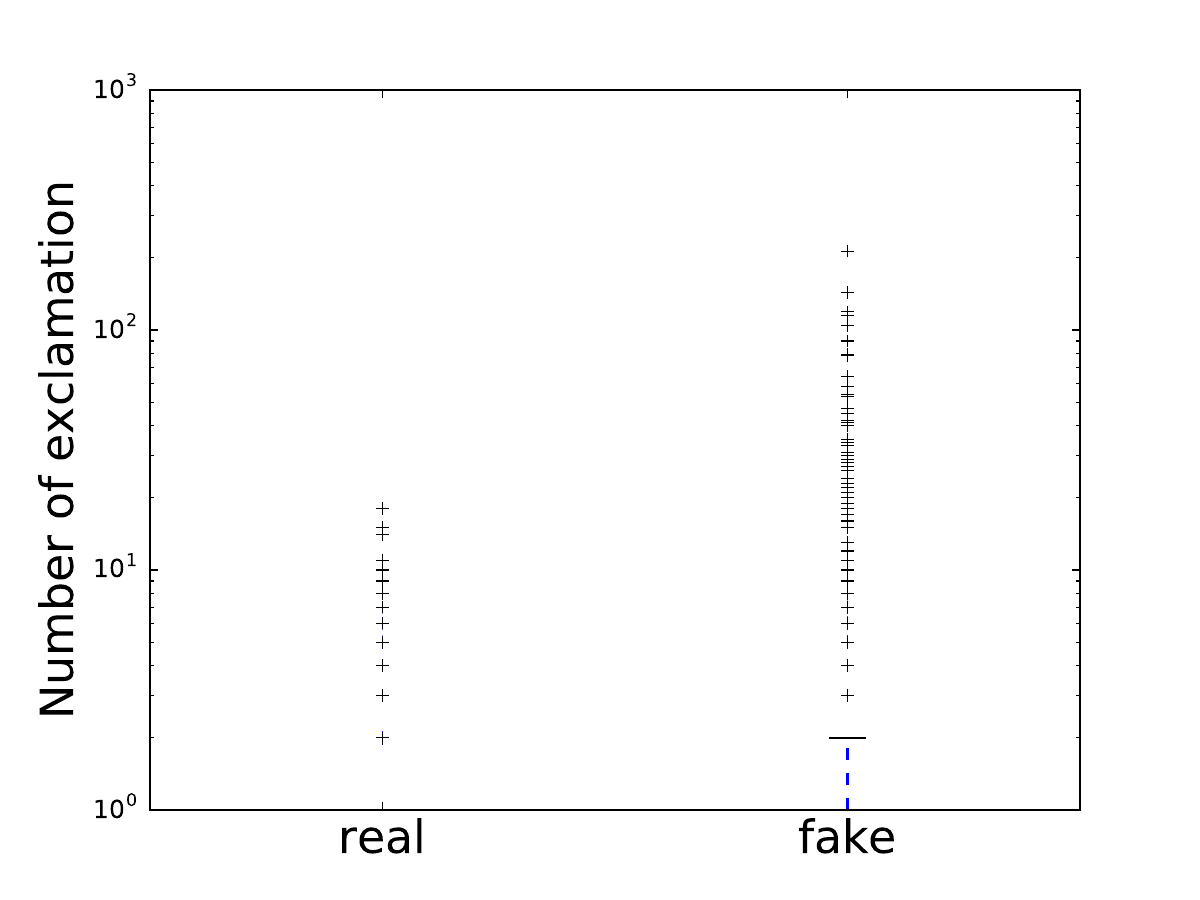}}
\subfigure[b][The exclusive words in news.]{\label{fig:exclusive}
\includegraphics[width=0.23\textwidth]{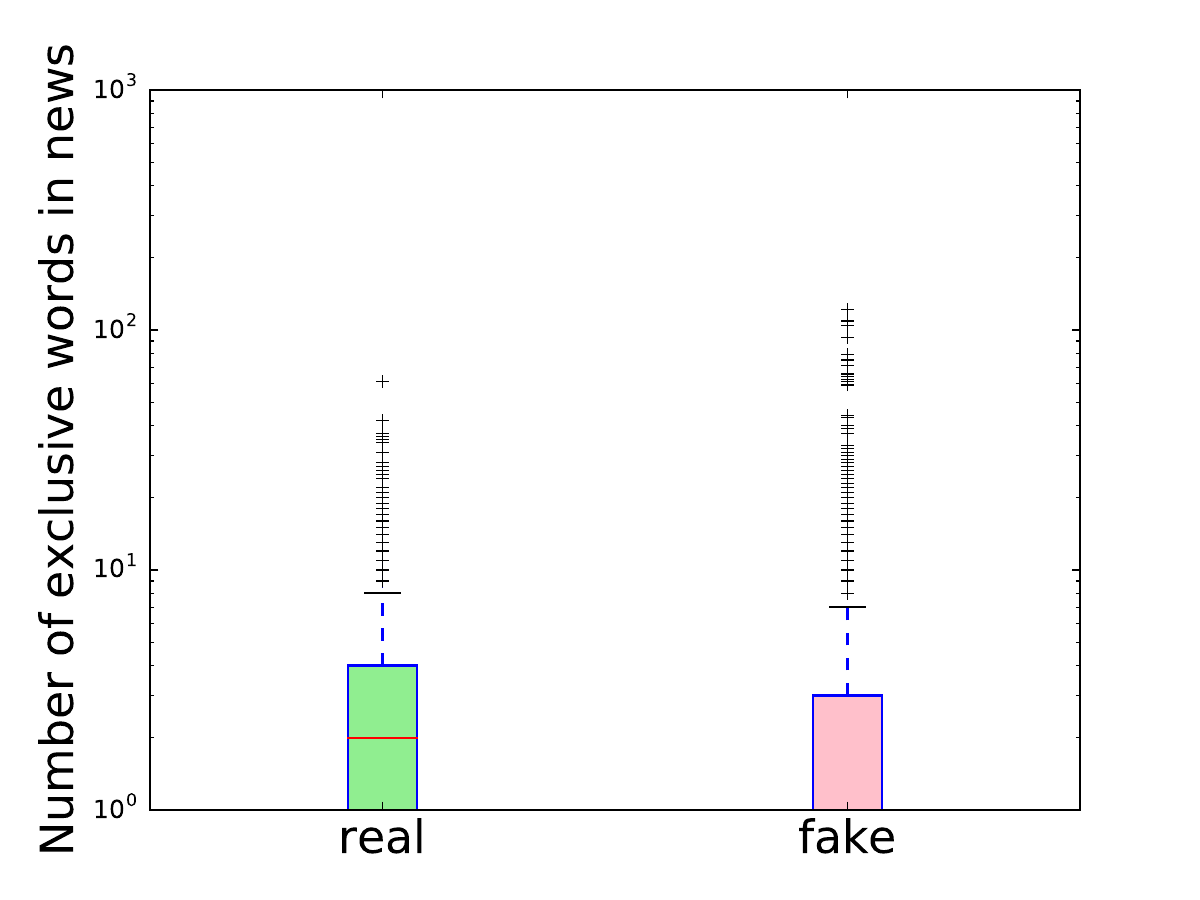}}
\subfigure[b][The negations in news.]{\label{fig:negations}
\includegraphics[width=0.23\textwidth]{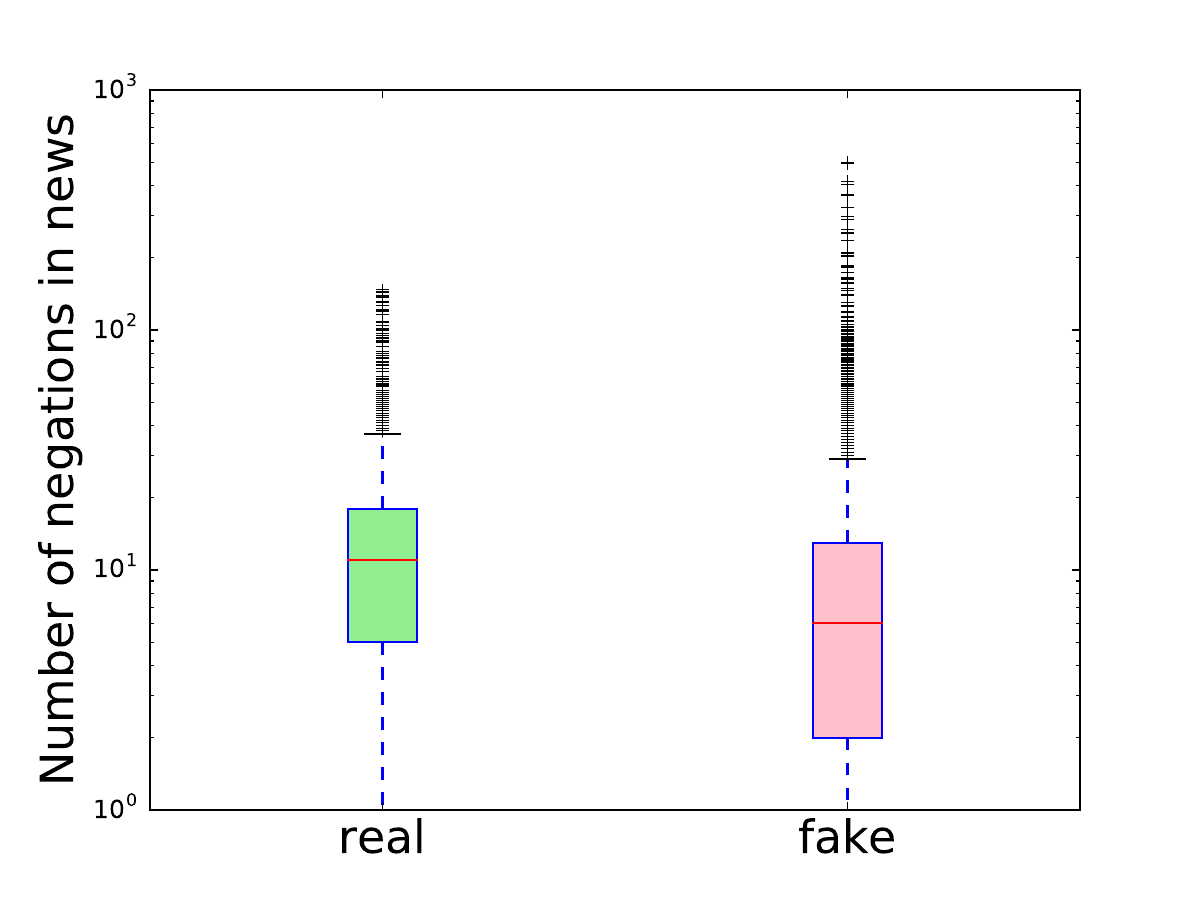}}
\subfigure[b][FPP: First-person pronoun.]{\label{fig:1st}
\includegraphics[width=0.23\textwidth]{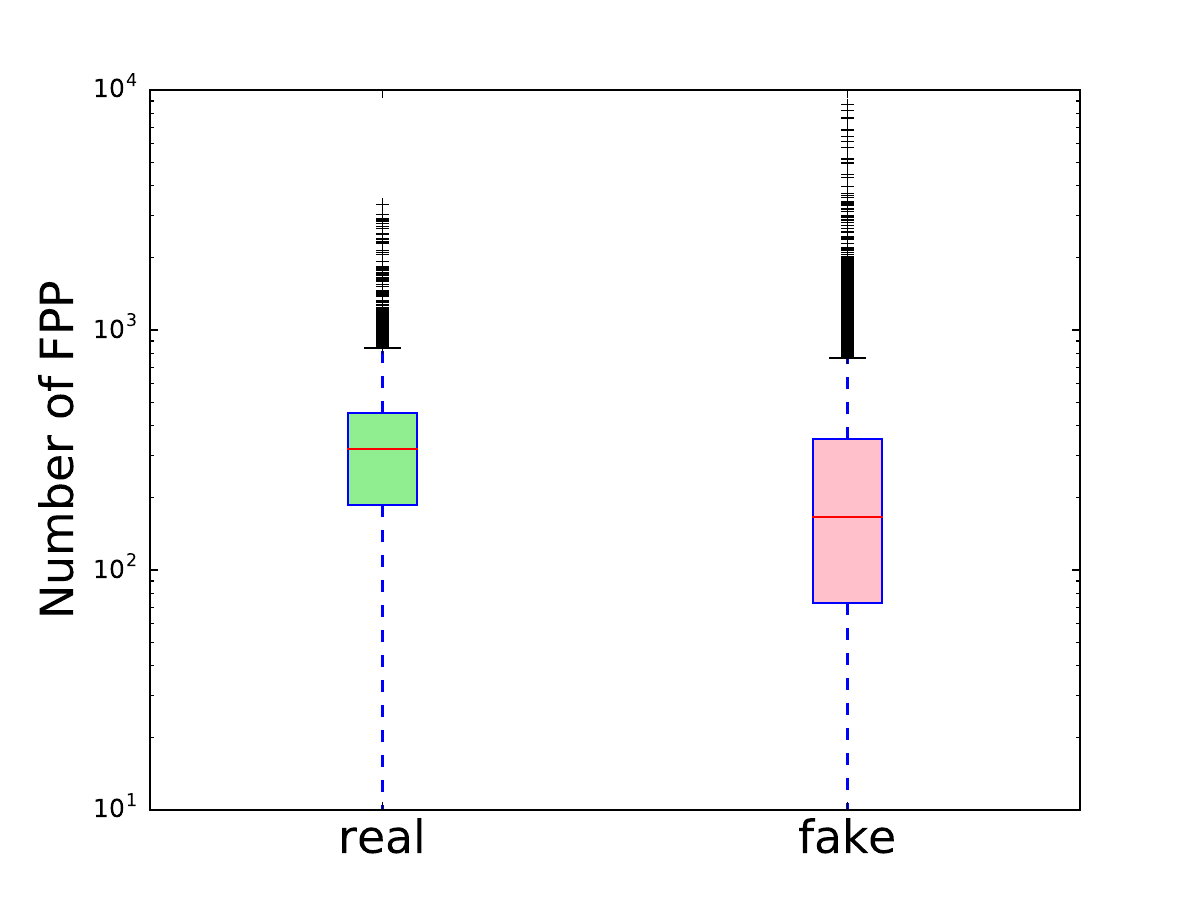}}
\subfigure[b][Motion verbs in news.]{\label{fig:motion_verbs}
\includegraphics[width=0.23\textwidth]{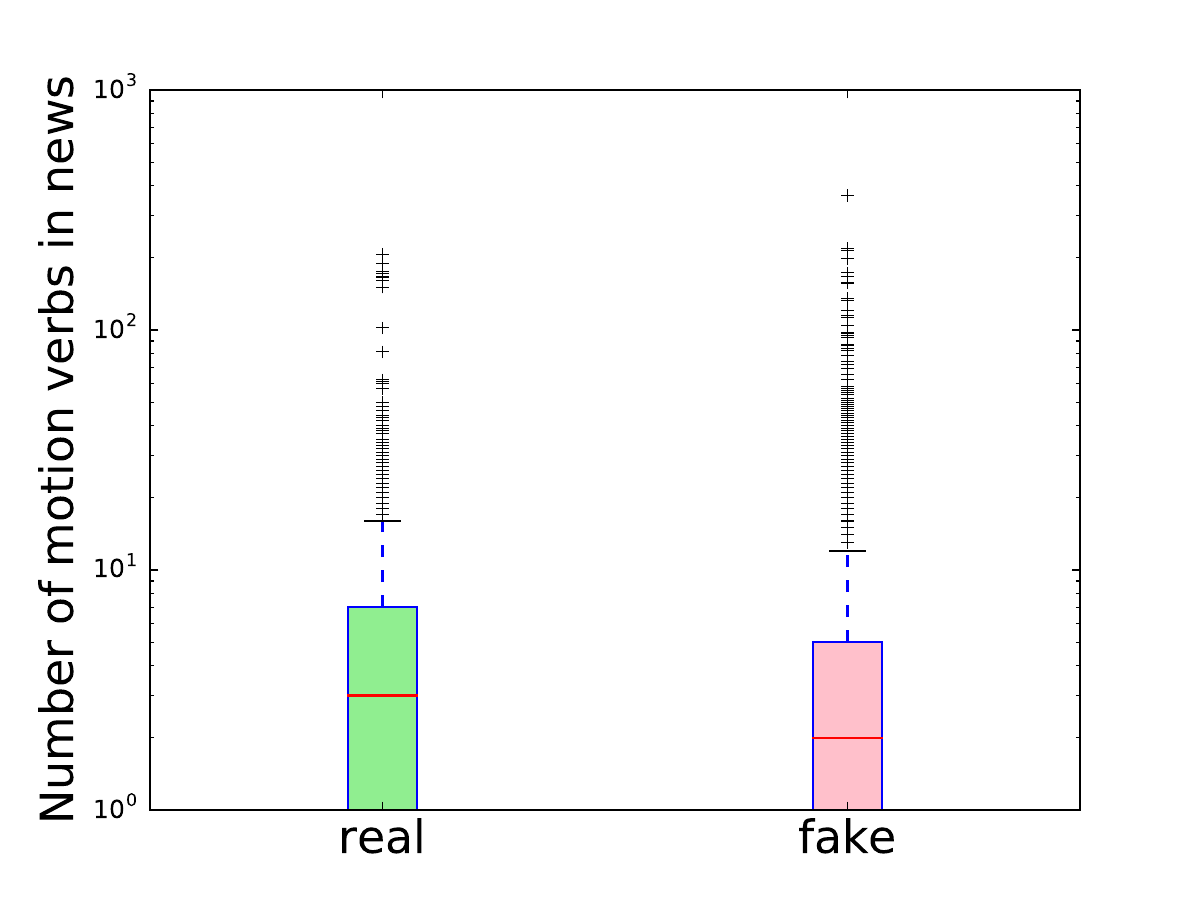}}
\caption{Analysis on the news text. }
\end{figure*}

\subsection{Text Analysis}
Let's take the word frequency \cite{kessler2017scattertext} in the titles as an example to demonstrate the differences between real and fake news in Fig. \ref{fig:word_frequency}. If the news has no title, we set the title as `notitle'. The frequently observed words in the title of fake news are \emph{notitle, IN, THE, CLINTON} and many meaningless numbers that represent special characters. We can have some interesting findings from the figure. Firstly, much fake news have no titles. These fake news are widely spread as the tweet with a few keywords and hyperlink of the news on social networks. Secondly, there are more capital characters in fake news. The purpose is to draw the readers' attention, while the real news contains less capital letters, which is written in a standard format. Thirdly, the real news contain more detailed descriptions. For example, names (\emph{Jeb Bush, Mitch McConnell}, etc.), and motion verbs (\emph{left, claim, debate and poll}, etc.). 

\subsubsection{Computational Linguistic}
\paragraph{Number of words and sentences}
Although liars have some control over the content of their stories, their underlying state of mind may leak out through the style of language used to tell the story. The same is true for the people who write the fake news. The data presented in the following paragraph provides some insight into the linguistic manifestations of this state of mind \cite{hancock2007lying}.

As shown in Fig. \ref{fig:words}, fake news has fewer words than real news on average. There are 4,360 words on average for real news, while the number is 3,943 for fake news. Besides, the number of words in fake news distributes over a wide range, which indicates that some fake news have very few words and some have plenty of words. The number of words is just a simple view to analyze the fake news. Besides, real news has more sentences than fake news on average. Real news has 84 sentences, while fake news has 69 sentences. Based on the above analysis, we can get the average number of words in a sentence for real and fake news, respectively. As shown in Fig. \ref{fig:avg_words}, the sentence of real news is shorter than that of fake news. Real news has 51.9 words on average in a sentence. However, the number is 57.1 for fake news. According to the box plot, the variance of the real news is much smaller than that of fake news. And this phenomenon appears in almost all the box plots. The reason is that the editor of real news must write the article under certain rules of the press. These rules include the length, word selection, no grammatical errors, etc. It indicates that most of the real news are written in a more standard and consistent way. However, most of the people who write fake news don't have to follow these rules.


\paragraph{Question mark, exclamation and capital letters}
According to the statistics on the news text, real news has fewer question marks than fake news, as shown in Fig. \ref{fig:question_mark}. The reasons may lie in that there are many rhetorical questions in fake news. These rhetorical questions are always used to emphasize the ideas consciously and intensify the sentiment. 

According to the analysis on the data, we find that both real and fake news have very few exclamations. However, the inner fence of fake news box plot is much larger than that of real news, as shown in Fig. \ref{fig:exclamation}. Exclamation can turn a simple indicative or declarative sentence into a strong command or reflect an emotional outburst. Hence, fake news is inclined to use the words with exclamations to fan specific emotions among the readers. 

 Capital letters are also analyzed in the real and fake news. The reason for the capitalization in news is to draw readers attention or emphasize the idea expressed by the writers. According to the statistic data, fake news have much more capital letters than real news. It indicates that fake news deceivers are good at using the capital letters to attract the attention of readers, draw them to read it and believe it. 

\paragraph{Cognitive perspective}
From the cognitive perspective, we investigate the exclusive words (e.g., `but', `without', `however') and negations (e.g.,, `no', `not' ) used in the news. Truth tellers use negations more frequently, as shown in Fig. \ref{fig:exclusive} and \ref{fig:negations}. The exclusive words in news have the similar phenomenon with the negations. The median of negations in fake news is much smaller than that of real news. The deceiver must be more specific and precise when they use exclusive words and negations, to lower the likelihood that being caught in a contradiction. Hence, they use fewer exclusive words and negations in writing.
For the truth teller, they can exactly discuss what happened and what didn't happen in that real news writer witnessed the event and knew all the details of the event. Specifically, individuals who use a higher number of “exclusive” words are generally healthier than those who do not use these words \cite{pennebaker1999linguistic}. 

\subsubsection{Psychology Perspective}
From the psychology perspective, we also investigate the use of first-person pronouns (e.g., I, we, my) in the real and fake news. Deceptive people often use language that minimizes references to themselves. A person who's lying tends not to use ``we” and ``I", and tend not to use person pronouns. Instead of saying ``I didn’t take your book,'' a liar might say ``That's not the kind of thing that anyone with integrity would do'' \cite{newman2003lying}. Similarly, as shown in Fig. \ref{fig:1st}, the result is the same with the point of view from the psychology perspective. On average, fake news has fewer first-person pronouns. The second-person pronouns (e.g., you, yours) and third-person pronouns (e.g., he, she, it) are also tallied up. We find that deceptive information can be characterized by the use of fewer first-person, fewer second-person and more third-person pronouns. Given space limitations, we just show the first-person pronouns figure. In addition, the deceivers avoid discussing the details of the news event. Hence, they use few motion verbs, as shown in Fig. \ref{fig:motion_verbs}.

\subsubsection{Lexical Diversity}
Lexical diversity is a measure of how many different words that are used in a text, while lexical density provides a measure of the proportion of lexical items (i.e. nouns, verbs, adjectives and some adverbs) in the text. The rich news has more diversity. According to the experimental results, the lexical diversity of real news is $2.2e\text{-}06$, which is larger than $1.76e\text{-}06$ for fake news.

\subsubsection{Sentiment Analysis}
\begin{figure}[ht]
\centering
\subfigure[b][The median sentiment values: positive and negative.]{\label{fig:median_pos_neg}
\includegraphics[width=0.23\textwidth]{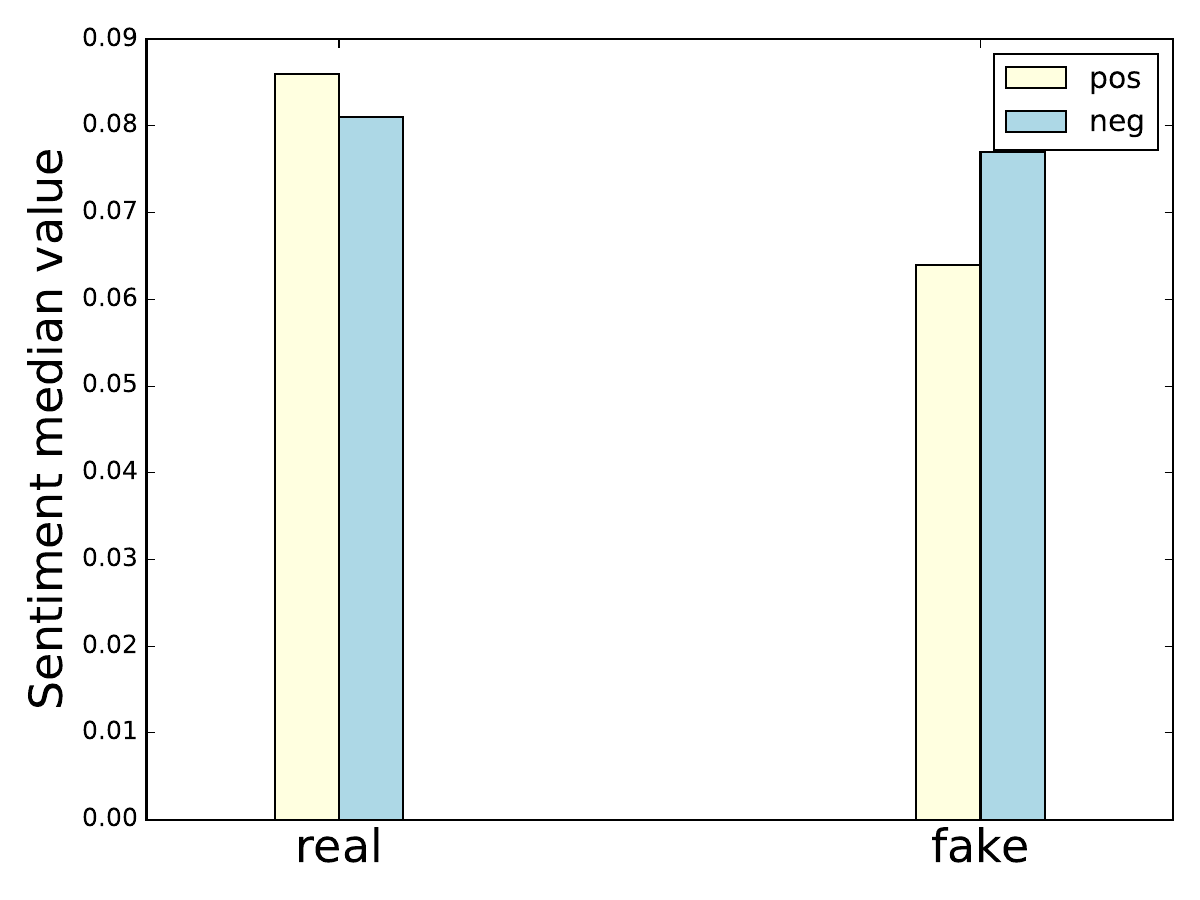}}
\subfigure[b][The standard deviation sentiment values: positive and negative.]{\label{fig:sd_pos_neg}
\includegraphics[width=0.23\textwidth]{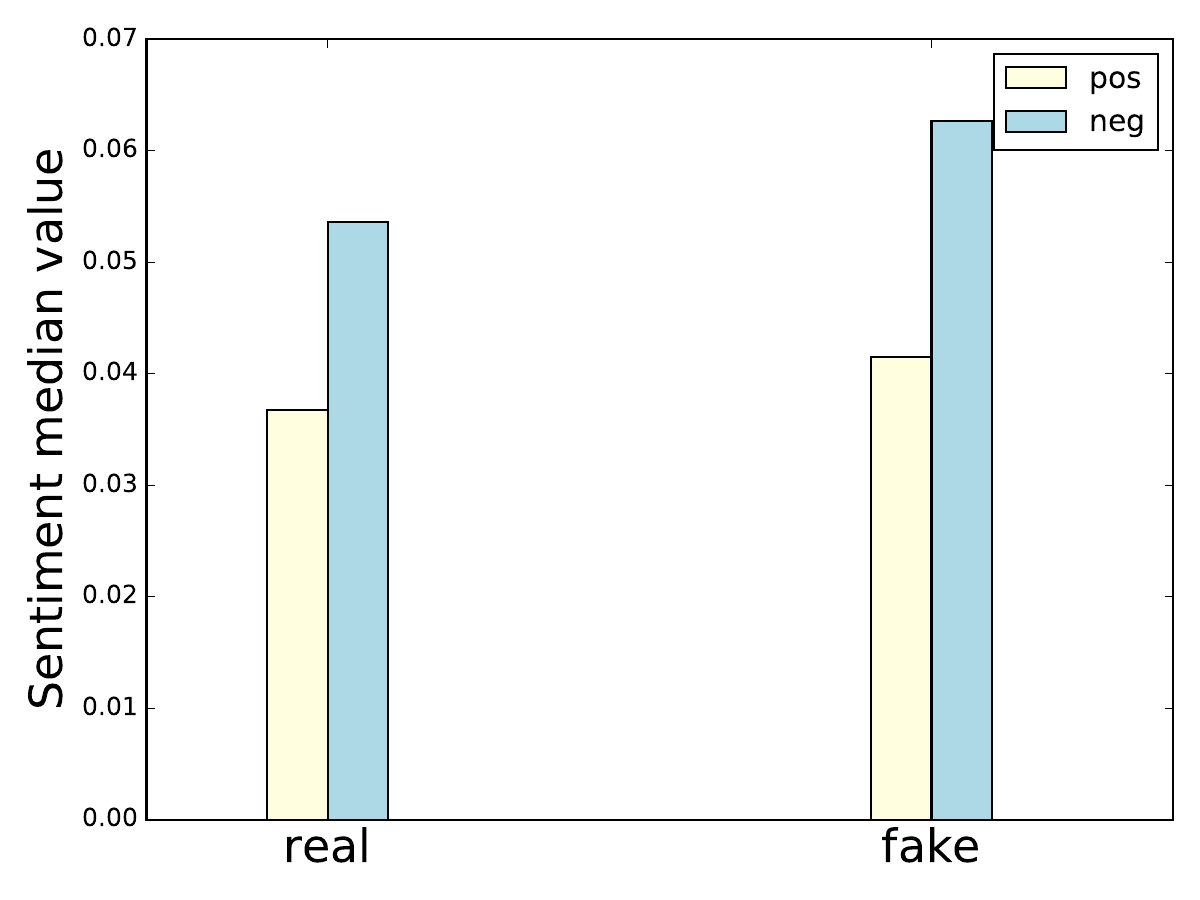}}
\caption{Sentiment analysis on real and fake news.}
\label{fig:sentiment}
\end{figure}

The sentiment \cite{liu2012sentiment} in the real and fake news is totally different. For real news, they are more positive than negative ones. The reason is that deceivers may feel guilty or they are not confident to the topic. Under the tension and guilt, the deceivers may have more negative emotion \cite{markowitz2016linguistic,pennebaker1999linguistic}. The experimental results agree with the above analysis in Fig. \ref{fig:sentiment}. The standard deviation of fake news on negative sentiment is also larger than that of real news, which indicates that some of the fake news have very strong negative sentiment.

\subsection{Image Analysis}
We also analyze the properties of images in the political news. According to some observations on the images in the fake news, we find that there are more faces in the real news. Some fake news have irrelevant images, such as animals and scenes. The experiment result is consistent with the above analysis. There are 0.366 faces on average in real news, while the number is 0.299 in fake news.
In addition, real news has a better resolution image than fake news.
The real news has $457 \times 277$ pixels on average, while the fake news has a resolution of $355\times 228$.

\section{Model -- the architecture}
In this section, we introduce the architecture of TI-CNN model in detail. Besides the explicit features, we innovatively utilize two parallel CNNs to extract latent features from both textual and visual information. And then explicit and latent features are projected into the same feature space to form new representations of texts and images. At last, we propose to fuse textual and visual representations together for fake news detection.

As shown in Fig. \ref{fig:architecture}, the overall model contains two major branches, i.e., text branch and image branch. For each branch, taking textual or visual data as inputs, explicit and latent features are extracted for final predictions. To demonstrate the theory of constructing the TI-CNN, we introduce the model by answering the following questions: 1) How to extract the latent features from text? 2) How to combine the explicit and latent features? 3) How to deal with the text and image features together? 4) How to design the model with fewer parameters? 5) How to train and accelerate the training process?


\renewcommand{\arraystretch}{0.7}

\begin{table}[htbp]  \centering  \caption{Symbols in this paper.}    \begin{tabular}{lll}    \toprule    \multicolumn{1}{c}{Parameter} & \multicolumn{1}{l}{Parameter Name } & \multicolumn{1}{c}{Dimension} \\    \midrule       
$\mathbf{X}^{Tl}_{i,j}$   & latent word vector $j$ in sample $i$      & $\mathbb{R}^k$ \\ 
$\mathbf{X}^{Tl}_{i,1:n}$  &  sentence for sample $i$   &  $\mathbb{R}^{n\times k}$ \\
$\mathbf{X}_i^{Te}$   & explicit text feature for sample $i$    & $\mathbb{R}^k$ \\ 
$\mathbf{X}_i^{Ie}$   & explicit image feature for sample $i$    & $\mathbb{R}^k$ \\ 
$\mathbf{X}_i^{Il}$   & latent image feature for sample $i$   & $\mathbb{R}^k$ \\ 
$\theta$  &  weight for the word     & $\mathbb{R}^{h \times k}$ \\        
$\mathbb{Y}$  & label of news    & $\mathbb{R}^{n\times 2}$ \\
       $w$  &  filter for texts     & $\mathbb{R}^{h \times k}$ \\        $b$ &    bias   &  $\mathbb{R}$ \\
       $\mathbf{c}$  &  feature map     & $\mathbb{R}^{n-h+1}$  \\        $\hat{c}$  & the maximum value in feature map &  $\mathbb{R}$ \\         $M$ &  number of maps     &  $\mathbb{R}$ \\         $M_i$ & the i-th filter for images &  $\mathbb{R}^{K_\alpha \times K_\beta}$ \\
            $\tau$     & the scores in tags of label       & $\mathbb{R}$ \\
            $T$         &  the number of tags in label     &  $\mathbb{R}$ \\         $s_{w}(\mathbb{X})_\tau$ &  the predicted probability &  $\mathbb{R}\in [0,1]$  \\    \bottomrule    \end{tabular}  \label{tab:symbols}%
            \end{table}%

\subsection{Text Branch}
For the text branch, we utilize two types of features: textual explicit features $\mathbf{X}^{Te}$ and textual latent features $\mathbf{X}^{Tl}$. The textual explicit features are derived from the statistics of the news text as we mentioned in the data analysis part, such as the length of the news, the number of sentences, question marks, exclamations and capital letters, etc. The statistics of a single news can be organized as a vector with fixed size. Then the vector is transformed by a fully connected layer to form a textual explicit features. 

\begin{figure*}[ht]
\begin{center}
 \includegraphics[width=0.9\textwidth]{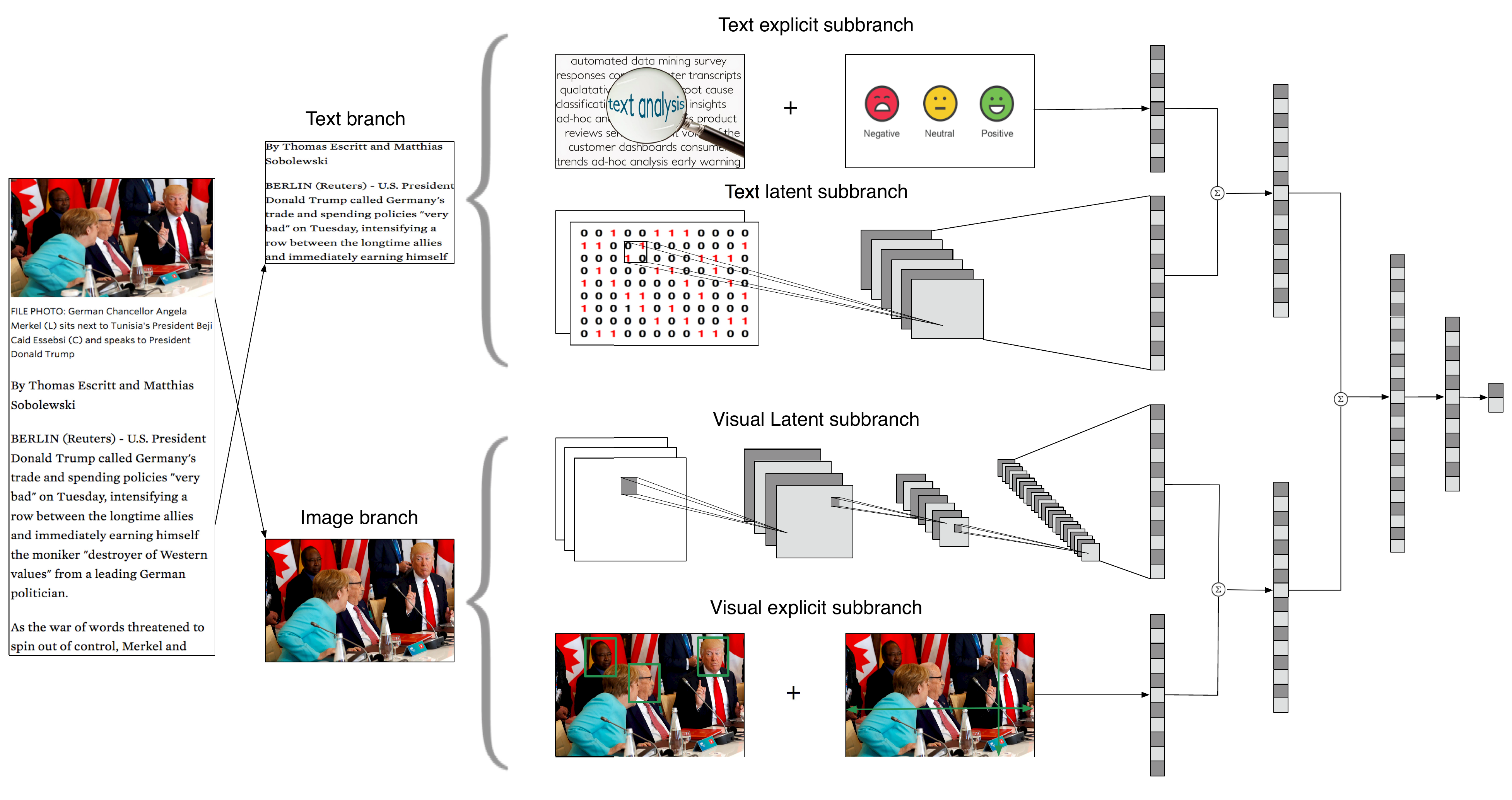}
\end{center}
\caption{The architecture of the model. The rectangles in the last 5 layers represent the hidden dense layers. The dropout, batch normalization and flatten layers are not drawn for brevity. The details of the structure are shown in Table \ref{tab:hyperparameter}.}
\label{fig:architecture}
\end{figure*} 

The textual latent features in the model are based on a variant of CNN. Although CNNs are mainly used in Computer Vision tasks, such as image classification \cite{krizhevsky2012imagenet} or object recognition \cite{ren2015faster}, CNN also show notable performances in many Natural Language Processing (NLP) tasks \cite{kim2014convolutional,zeng2014relation}. With the convolutional approach, the neural network can produce local features around each word of the adjacent word and then combines them using a max operation to create a fixed-sized word-level embedding, as shown in Fig. \ref{fig:architecture}. Therefore, we employ CNN to model textual latent features for fake news detection. Let the $j$-th word in the news $i$ denote as $\mathbf{x}_{i,j} \in \mathbb{R}^{k}$, which is a k-dimensional word embedding vector. Suppose the maximum length of the news is $n$, s.t., the news have less than $n$ words can be padded as a sequence with length $n$. Hence, the overall news can be written as 
\begin{equation}
\mathbf{X}_{i,1:n}^{Tl}=\mathbf{x}_{i,1} \oplus \mathbf{x}_{i,1} \oplus \mathbf{x}_{i,2} \oplus ... \oplus \mathbf{x}_{i,n}.
\end{equation}

It means that the news $\mathbf{X}_{i,1:n}^{Tl}$ is concatenated by each word. In this case, each news can be represented as a matrix. Then we use convolutional filters $w \in \mathbb{R}^{h\times k}$ to construct the new features. For instance, a window of words $\mathbf{X}_{i,j:j+h-1}^{Tl}$ can produce a feature $c_i$ as follows:
\begin{equation}
c_i=f(\mathbf{w} \cdot \mathbf{X}_{i,j:j+h-1}^{Tl}+b),
\end{equation}
where the $b\in R$ is the bias, and $\cdot$ is the convolutional operation. $f$ is the non-linear transformation, such as the sigmoid and tagent function. A feature map is generated from the filter by going through all the possible window of words in the news.
\begin{equation}
\mathbf{c}=[c_1,c_2,...,c_{n-h+1}],
\end{equation}
where $\mathbf{c}\in \mathbb{R}^{n-h+1}$. A max-pooling layer \cite{nagi2011max} is applied to take the maximum in the feature map $\mathbf{c}$. The maximum value is denoted as $\hat{c}=max\{\mathbf{c}\}$. The max-pooling layer can greatly improve the robustness of the model by reserving the most important convolutional results for fake news detection. The pooling results are fed into a fully connected layer to obtain our final textual latent features for predicting news labels.


\subsection{Image Branch}
 Similar to the text branch, we use two types of features: visual explicit features $\mathbf{X}^{Ie}$ and visual latent features $\mathbf{X}^{Il}$. As shown in Fig. \ref{fig:architecture}, in order to obtain the visual explicit features, we firstly extract the resolution of an image and the number of faces in the image to form a feature vector. And then, we transform the vector into our visual explicit feature with a fully connected layer.

Although visual explicit features can convey information of images contained in the news, it is hand-crafted features and not data-driven. To directly learn from the raw images contained in the news to derive more powerful features, we employ another CNN to learn from images in the news.

\subsubsection{Convolutional layer}
In the convolutional layer, filters are replicated across entire visual field and share the same parameterisation forming a feature map. In this case, the network have a nice property of translation-invariant. Suppose the convolutional layer has $M$ maps of size $(M_\alpha,M_\beta)$. A filter $(K_\alpha,K_\beta)$ is shifted over all the regions of the images. Hence the size of the output map is as follows:
\begin{equation}
M_{\alpha}^{n}=M_{\alpha}^{n-1} - K_{\alpha}^{n}+1
\end{equation}
\begin{equation}
M_{\beta}^{n}=M_{\beta}^{n-1} - K_{\beta}^{n}+1
\end{equation}

\subsubsection{Max-pooling layer}
A max-pooling layer \cite{nagi2011max} is connected to the convolutional layer. Then we apply maximum activation over the rectagular filters $(K_\alpha,K_\beta)$ to the output of max-pooling layer.  Max-pooling enables position invariance over larger local regions and downsamples the input image by a factor of $K_\alpha$ and $K_\beta$ along each direction, which can make the model select invariant features, converge faster and improve the generalization significantly. A theoretical analysis of feature pooling in general and max-pooling in particular is given by \cite{boureau2010theoretical}.
\begin{table*}[ht]
\centering
\caption{Two fake news correspond to the Fig. \ref{fig:amish} and \ref{fig:hillary}.}
\label{tab:case_study}
\resizebox{\textwidth}{!}{%
\begin{tabular}{|>{\arraybackslash}M{2.5cm}|>{\arraybackslash}m{14cm}|>{\centering\arraybackslash}M{0.5cm}|}
\hline
\multicolumn{1}{|>{\centering\arraybackslash}p{23mm}|}{\thead{ \textbf{Title}}} 
    & \multicolumn{1}{>{\centering\arraybackslash}p{8cm}|}{\thead{ \textbf{\ \ \ \ \ \ \ \ \ \ \ \ \ \ \ \ \ \ \ \ \ \ \ \ \ \ \ \ \ \ \ \ \ \ \ \ \ \ \ \ \ \ \ \ \ News text}}} 
    & \multicolumn{1}{>{\centering\arraybackslash}p{0.2cm}|}{\textbf{Type}} \\ \hline
The Amish Brotherhood have endorsed Donald Trump for president. & The Amish, who are direct descendants of the protestant reformation sect known as the Anabaptists, have typically stayed out of politics in the past. As a general rule, they don't vote, serve in the military, or engage in any other displays of patriotism. This year, however, the AAB has said that it is imperative that they get involved in the democratic process. & Fake \\ \hline
Wikileaks Gives Hillary An Ultimatum: QUIT, Or We Dump Something Life-Destroying & On Sunday, Wikileaks gave Hillary Clinton less than a 24-hour window to drop out of the race or they will dump something that will destroy her “completely.”Recently, Julian Assange confirmed that WikiLeaks was not working with the Russian government, but in their pursuit of justice, they are obligated to release anything that they can to bring light to a corrupt system – and who could possibly be more corrupt than Crooked Hillary? & Fake \\ \hline
\end{tabular}%
}
\end{table*}
\subsection{Rectified Linear Neuron}
The sigmoid and tanh activation functions may cause the gradient explode or vanishing problem \cite{pascanu2013difficulty} in convolutional neural networks. Hence, we add the ReLU activation to the image branch to remede the problem of gradient vanishing.
\begin{equation}
    y = max(0,\sum_{i=1}^{k}x_{i}\theta_{i}+b)
\end{equation}
ReLUs can also improve neural networks by speeding up training. The gradient computation is very simple (either 0 or 1 depending on the sign of $x$). Also, the computational step of a ReLU is easy: any negative elements are set to 0.0 -- no exponentials, no multiplication or division operations.

Logistic and hyperbolic tangent networks suffer from the vanishing gradient problem, where the gradient essentially becomes 0 after a certain amount of training (because of the two horizontal asymptotes) and stops all learning in that section of the network. ReLU units are only 0 gradient on one side, which is empirically superior. 

\subsection{Regularization}
As shown in Table \ref{tab:hyperparameter}, we empoly dropout \cite{srivastava2014dropout} as well as $l_2$-norms to prevent overfitting. Dropout is to set some of the elements in weight vectors as zero with a probability $p$ of the hidden units during the forward and backward propagation. For instance, we have a dense layer $z=[z_1,...,z_m]$, and $r$ is a vector where all the elements are zero. When we start to train the model, the dropout is to set some of the elements of $r$ as 1 with probability as $p$. Suppose the output of dense layer is $y$. Then the dropout operation can be formulated as 
\begin{equation}
    y=\theta \cdot (z\circ r)+b,
\end{equation}
where $\theta$ is the weight vector. $\circ$ is the element-wise multiplication operator.
When we start to test the performance on the test dataset, the deleted neurons are back. The deleted weight are scaled by $p$ such that $\hat{\theta} =p\theta$. The $\hat{\theta}$ is used to predict the test samples. The above procedure is implemented iteratively, which greatly improve the generalization ability of the model. We also use early stopping \cite{prechelt1998automatic} to avoid overfitting. It can also be considered a type of regularization method (like L1/L2 weight decay and dropout).

\subsection{Network Training}
We train our neural network by minimizing the negative likelihood on the training dataset $D$. To identify the label of a news $\mathbb{X}$, the network with parameter $\theta$ computes a value $s_{w}(x)_{\tau}$. Then a sigmoid function is used over all the scores of tags $\tau \in T$ to transform the value into the conditional probability distribution of labels:
\begin{equation}
p(\tau|\mathbb{X},\theta)=\frac{e^{s_{\theta}(\mathbb{X})_{\tau}}}{\sum_{\forall i\in T}{e^{s_{\theta}(\mathbb{X})_i}}}
\label{equ:equation}
\end{equation}

The negative log likelihood of Equation \ref{equ:equation} is 
\begin{equation}
E(W)=-lnp(\tau|\mathbb{X},\theta)=s_{\theta}(\mathbb{X})_{\tau}-log\left (\sum_{\forall i\in T}e^{s_{\theta}(\mathbb{X})_{\tau}}\right)
\end{equation}

We use the RMSprop \cite{hinton2012neural} to minimize the loss function with respect to parameter $\theta$:
\begin{equation}
\theta -> \sum_{(\mathbb{X},\mathbb{Y})\in D}{-log\ p(\mathbb{Y}|\mathbb{X},\theta)}
\end{equation}

where $\mathbb{X}$ is the input data, and $\mathbb{Y}$ is the label of the news. We naturally choose back-propagation algorithm \cite{hecht1988theory} to compute the gradients of the network structure. With the fine-tuned parameters, the loss converges to a good local minimum in a few epochs.




\section{Experiments}

\subsection{Case study}
\label{ssec:case_study}
A case study of the fake news is given in this section. The two fake news in Table \ref{tab:case_study} correspond to the Fig. \ref{fig:amish} and \ref{fig:hillary}. The first fake news is an article reporting that `the American Amish Brotherhood endorsed Donald Trump for President'. However, the website is a fake CNN page. The image in the fake news can be easily searched online, and it is not very relevant with the news texts\footnote{http://cnn.com.de/news/amish-commit-vote-donald-trump-now-lock-presidency/}. For the second fake news -- `Wikileaks gave Hillary Clinton less than a 24-hour window to drop out of the race', it is actually not from Wikileaks. Besides, the composite image \footnote{http://thelastlineofdefense.org/wikileaks-gives-hillary-an-ultimatum-quit-or-we-dump-something-life-destroying/} in the news is low quality.


\subsection{Experimental Setup}
We use 80\% of the data for training, 10\% of the data for validation and 10\% of the data for testing. All the experiments are run at least 10 times separately. The textual explicit subbranch and visual explicit subbranch are connected with a dense layer. The parameters in these subbranches can be learned easily by the back-propagation algorithm. Thus, most of the parameters, which need to be tuned, exist in the textual latent subbranch and visual latent subbranch. The parameters are set as follows.

\subsubsection{Text branch}
For the textual latent subbranch, the embedding dimension of the word2vec is set to 100. The details of how to select the parameters are demonstrated in the sensitivity analysis section. The context of the word2vec is set to 10 words. The filter size in the convolutional neural network is $(3,3)$. There are 10 filters in all. Two dropouts are adopted to improve the model's generalization ability. 
For the textual explicit subbranch, we add a dense layer with 100 neurons first, and then add a batch normalization layer to normalize the activations of the previous layer at each batch, i.e. applies a transformation that maintains the mean activation close to 0 and the activation standard deviation close to 1. 
The outputs of textual explicit subbranch and textual latent feature subbranch are combined by summing the outputs up.
\renewcommand{\arraystretch}{0.55}
\renewcommand{\arraystretch}{0.55}

\begin{table}[]
\centering
\caption{Models specifications. BN: Batch Normalization, ReLU: Rectified Linear Activation Function, Conv: Convolutional Layer on 2D data, Conv1D: Convolutional Layer on 1D data, Dense: Dense Layer, Emb: Embedding layer, MaxPo: Max-Pooling on 2D data, MaxPo1D: Max-Pooling on 1D data. There are two kinds of dropout layers, i.e., $D=(D_\alpha,D_\beta)$, where $D_\alpha = 0.5$ and $D_\beta = 0.8$.}
\label{tab:hyperparameter}
\begin{tabular}{|c|c|c|c|}
\hline
\multicolumn{2}{|c|}{Text Branch}                                                                                       & \multicolumn{2}{c|}{Image Branch}                                                                                     \\ \hline
\begin{tabular}[c]{@{}c@{}}Textual \\ Explicit\end{tabular} & \begin{tabular}[c]{@{}c@{}}Textual \\ Latent\end{tabular} & \begin{tabular}[c]{@{}c@{}}Visual \\ Latent\end{tabular} & \begin{tabular}[c]{@{}c@{}}Visual \\ Explicit\end{tabular} \\ \hline
\multirow{3}{*}{Input 31$\times$1}                                 & Emb 1000$\times$100                                             & Input 50$\times$50$\times$3                                            & \multirow{3}{*}{Input 4$\times$1}                                 \\ \cline{2-3}
                                                            & \multirow{2}{*}{Dropout $D_\alpha$}                          & (2$\times$2) Conv(32)                                             &                                                            \\ \cline{3-3}
                                                            &                                                           & ReLU                                                     &                                                            \\ \hline
\multirow{4}{*}{Dense 128}                                  & Emb 1000$\times$100                                              & Dropout $D_\beta$                                           & \multirow{4}{*}{Dense 128}                                 \\ \cline{2-3}
                                                            & (3,3) Conv1D(10)                                          & (2$\times$2) Maxpo                                              &                                                            \\ \cline{2-3}
                                                            & 2 MaxPo1D                                                 & (2$\times$2) Conv(32)                                             &                                                            \\ \cline{2-3}
                                                            & Flatten                                                   & ReLU                                                     &                                                            \\ \hline
\multirow{6}{*}{BN}                                         & \multirow{2}{*}{Dense 128}                                &Dropout $D_\beta$                                           & \multirow{6}{*}{BN}                                        \\ \cline{3-3}
                                                            &                                                           & (2$\times$2) Maxpo                                              &                                                            \\ \cline{2-3}
                                                            & \multirow{2}{*}{BN}                                       & (2$\times$2) Conv(32)                                             &                                                            \\ \cline{3-3}
                                                            &                                                           & ReLU                                                     &                                                            \\ \cline{2-3}
                                                            & \multirow{2}{*}{ReLU}                                     &Dropout $D_\beta$                                          &                                                            \\ \cline{3-3}
                                                            &                                                           & (2$\times$2) Maxpo                                              &                                                            \\ \hline
\multirow{4}{*}{ReLU}                                       & \multirow{4}{*}{Dropout $D_\beta$ }                          & Flatten                                                  & \multirow{4}{*}{ReLU}                                      \\ \cline{3-3}
                                                            &                                                           & Dense 128                                                &                                                            \\ \cline{3-3}
                                                            &                                                           & BN                                                       &                                                            \\ \cline{3-3}
                                                            &                                                           & RelU                                                     &                                                            \\ \hline
\multicolumn{2}{|c|}{Merge}                                                                                             & \multicolumn{2}{c|}{Merge}                                                                                            \\ \hline
\multicolumn{4}{|c|}{Merge}                                                                                                                                                                                                                     \\ \hline
\multicolumn{4}{|c|}{ReLU}                                                                                                                                                                                                                      \\ \hline
\multicolumn{4}{|c|}{Dense 128}                                                                                                                                                                                                                 \\ \hline
\multicolumn{4}{|c|}{BN}                                                                                                                                                                                                                        \\ \hline
\multicolumn{4}{|c|}{Sigmoid}                                                                                                                                                                                                                   \\ \hline
\end{tabular}
\end{table}

\subsubsection{Image branch}
For the visual latent subbranch, all the images are reshaped as size $(50\times 50)$. Three convolutional layers are added to the network hierarchically. The filters size is set to $(3,3)$, and there are 32 filters for each convolutional layer followed by a ReLU activation layer. A maxpooling layer with pool size $(2,2)$ is connected to each convolutional layer to reduce the probability to be over-fitting. Finally, a flatten, batch normalization and activation layer is added to the model to extract the latent features from the images. For the explicit image feature subbranch, the input of the explicit features is connected to the dense layer with 100 neurons. And then a batch normalization and activation layer are added. The outputs of image convolutional neural network and explicit image feature subbranch are combined by summing the outputs up. We concatenate the outputs of text and image branch. An activation layer and dense layer are transforming the output into two dimensions. The labels of the news are given by the last sigmoid activation layer. In Table \ref{tab:hyperparameter}, we show the parameter settings in the TI-CNN model. The total number of parameters is 7,509,980, and the number of trainable parameters is 7,509,176.

\subsection{Experimental Results}
We compare our model with several competitive baseline methods in Table \ref{tab:baseline}. With image information only, the model cannot identify the fake news well. It indicates that image information is insufficient to identify the fake news. With text information, traditional machine learning method --- logistic regression \cite{hosmer2013applied} is employed to detect the fake news. However, logistic regression fails to identify the fake news using the text information. The reason is that the hyperplane is linear, while the raw data is linearly inseparable. GRU \cite{chung2014empirical} and Long short-term memory \cite{hochreiter1997long} with text information are inefficient with very long sequences, and the model with 1000 input length performs worse. Hence, we take the input length 400 as the baseline method. With text and image information, TI-CNN outperforms all the baseline methods significantly. 

\renewcommand{\arraystretch}{1}
\begin{table}[h!tbp]
  \centering
  \caption{The experimental results on many baseline methods. The number after the name of the model is the maximum input length for textual information. For those news text less than 1,000 words, we padded the sequence with $0$.}
    \begin{tabular}{cccc}
    \toprule
    \thead{ \textbf{Method}} & \thead{  \textbf{Precision}} & \thead{ \textbf{Recall}} & \thead{ \textbf{F1-measure}} \\
    \hline
        \textbf{CNN-image} & 0.5387  & 0.4215  & 0.4729  \\
    \textbf{LR-text-1000} & 0.5703  & 0.4114  & 0.4780  \\
    \textbf{CNN-text-1000} & 0.8722  & 0.9079  & 0.8897  \\
    \textbf{LSTM-text-400} & 0.9146  & 0.8704  & 0.8920  \\
    \textbf{GRU-text-400}   & 0.8875  & 0.8643  & 0.8758  \\
    \textbf{TI-CNN-1000} & \textbf{\ 0.9220 } & \textbf{\ 0.9277 } & \textbf{\ 0.9210 } \\
    \bottomrule
    \end{tabular}%
  \label{tab:baseline}%
\end{table}%

\subsection{Sensitivity Analysis}
In this section, we study the effectiveness of several parameters in the proposed model: the word embedding dimensions, batch size, the hidden layer dimensions, the dropout probability and filter size.
\begin{figure*}[ht]
\subfigure[b][Word embedding dimension and F1-measure.]{\label{fig:word_embedding}
\includegraphics[width=0.3\textwidth]{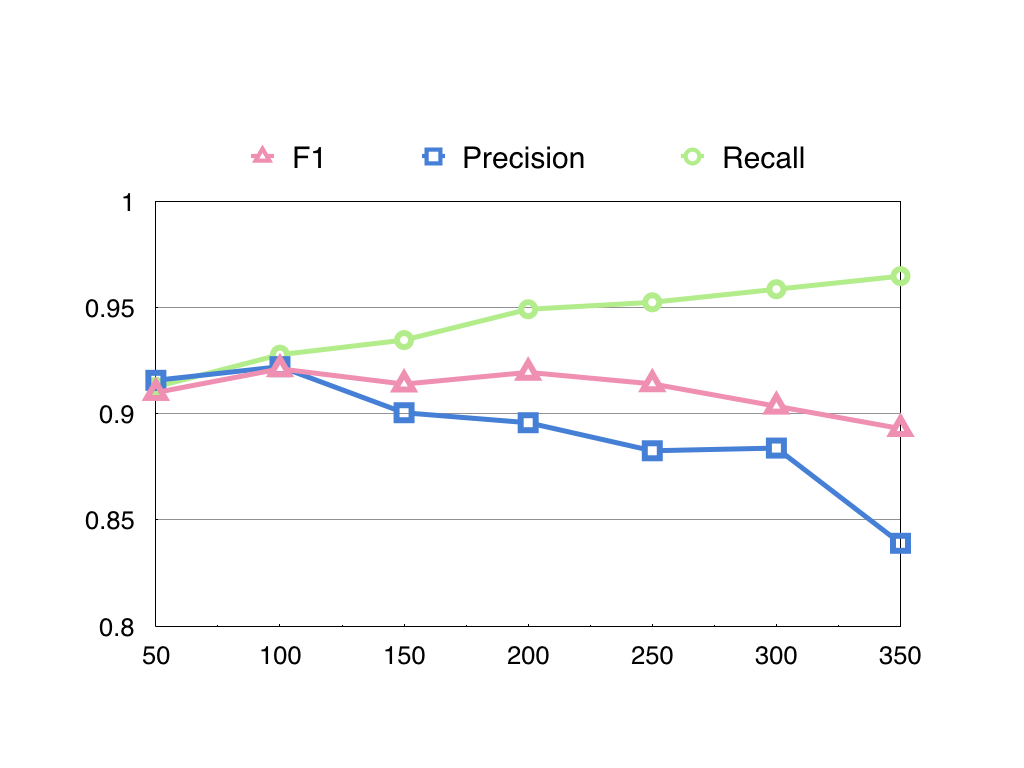}}
\subfigure[b][Batch size and F1-measure.]{\label{fig:batch_size}
\includegraphics[width=0.3\textwidth]{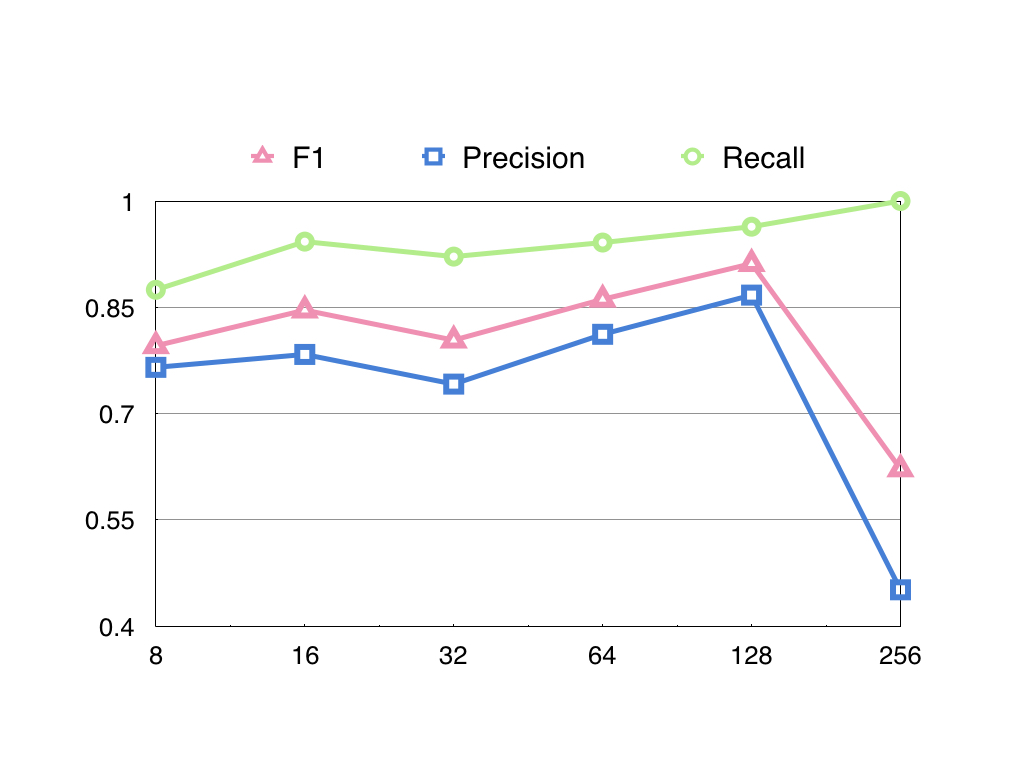}}
\subfigure[b][Hidden layer dimension and F1-measure.]{\label{fig:hidden_dimension}
\includegraphics[width=0.3\textwidth]{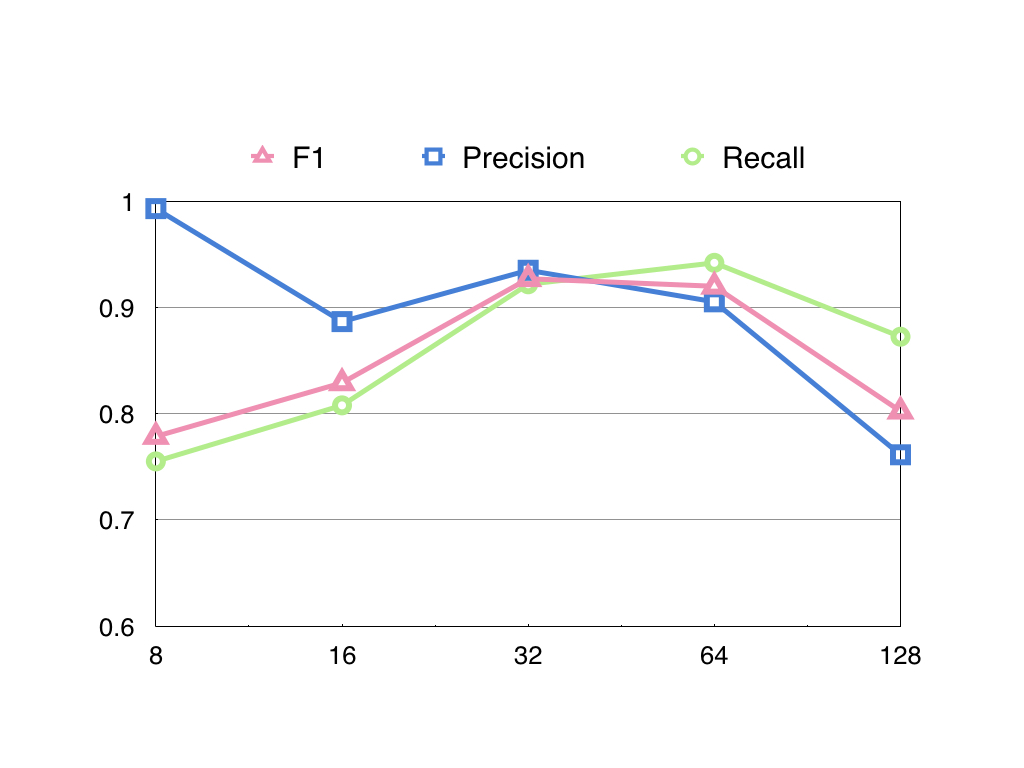}}
\caption{Word embedding dimension, batch size and the performance of the model.}
\label{fig:word_embedding_and_batch_size}
\end{figure*}

\paragraph{word embedding dimensions}
In the text branch, we exploit a 3 layer neural network to learn the word embedding. The learned word vector can be defined as a vector with different dimensions, i.e., from 50 to 350. In Fig. \ref{fig:word_embedding}, we plot the relation between the word embedding dimensions and the performance of the model. As shown in figure \ref{fig:word_embedding}, we find that the precision, recall and f1-measure increase as the word embedding dimension goes up from 50 to 100. However, the precision and recall decrease from 100 to 350. The recall of the model is growing all the time with the increase of the word embedding dimension. We select 100 as the word embedding dimension in that the precision, recall and f1-measure are balanced. For fake news detection in real world applications, the model with high recall is also a good choice. The reason is that publishers can use high recall model to collect all the suspected fake news at the beginning, and then the fake news can be identified by manual inspection.

\begin{figure}[h!t]
\subfigure[b][Dropout probabilities ($D_\alpha, D_\beta$) and the performance of the model.]{\label{fig:drop_out}
\includegraphics[width=0.4\textwidth]{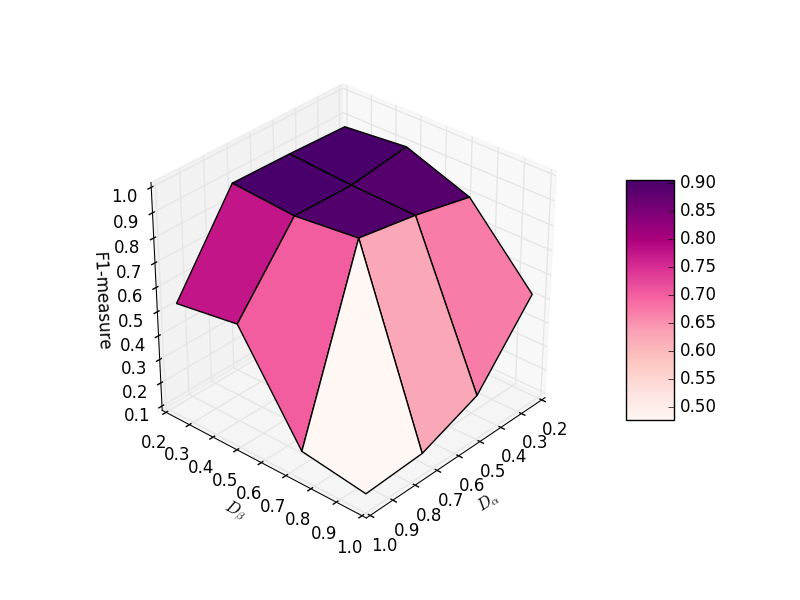}}
\subfigure[b][Filter size and the performance of the model.]{\label{fig:filter_size}
\includegraphics[width=0.4\textwidth]{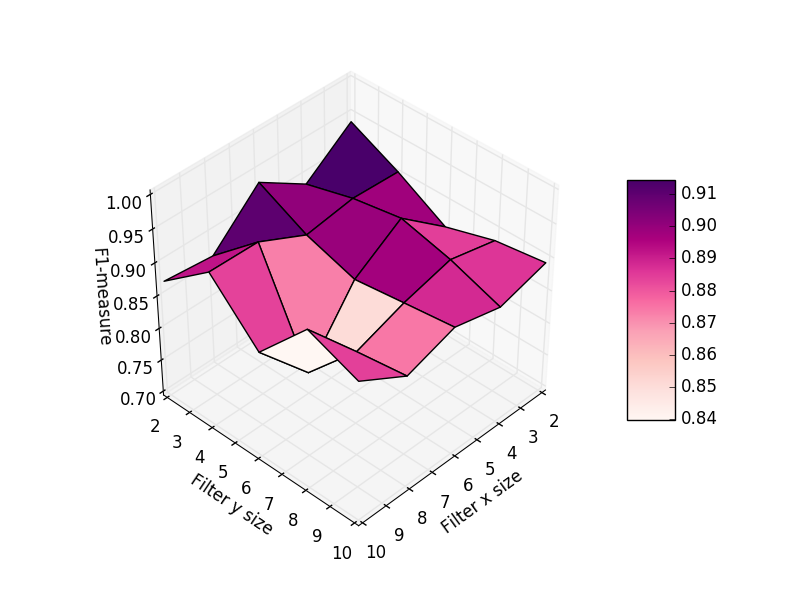}}
\caption{Dropout probabilities ($D_\alpha, D_\beta$), filter size and the performance of the model.}
\label{fig:word_embedding_and_batch_size}
\end{figure}

\paragraph{batch size}
Batch size defines the number of samples that going to be propagated through the network. The higher the batch size, the more memory space the program will need. The lower the batch size, the less time the training process will take. The relation between batch size and the performance of the model is shown in Fig. \ref{fig:batch_size}. The best choice for batch size is 32 and 64. The F1 measure goes up from batch size 8 to 32 first, and then drops when the batch size increases from 32 to 128. For batch size 8, it takes 32 seconds to train the data on each epoch. For batch size 128, it costs more than 10 minutes to train the model on each epoch.


\paragraph{hidden layer dimension}
As shown in Fig. \ref{fig:architecture}, there are many hidden dense layers in the model. 
Deciding the number of neurons in the hidden layers is a very important part of deciding the overall neural network architecture. Though these layers do not directly interact with the external environment, they have a tremendous influence on the final output. Using too few neurons in the hidden layers will result in underfitting. Using too many neurons in the hidden layers can also result in several problems. Some compromise must be reached between too many and too few neurons in the hidden layers. As shown in Fig. \ref{fig:hidden_dimension}, we find that 128 is the best choice for hidden layer dimension. The performance firstly goes up with the increase of the hidden layer dimension from 8 to 128. However, the dimension of the hidden layer reaches 256, the performance of the model drops due to overfitting.



\paragraph{Dropout probability and filter size}
We analyze the dropout probabilities, as shown in Table \ref{tab:hyperparameter}. $D_\alpha$ in Fig. \ref{fig:drop_out} is the dropout layer connected to the text embedding layer, while $D_\beta$ is used in both text and image branches. We use the grid search to choose the dropout probabilities. The model performs well when the $D_\alpha$ in the range [0.1,0.5] and the $D_\beta$ in range [0.1,0.8]. In this paper, we set the dropout probabilities as (0.5,0.8), which can improve the model's generalization ability and accelerate the training process. 

The filter size of a 1-dimension convolutional neural network layer in the textual latent subbranch is also a key factor in identifying the performance of the model. According to the paper \cite{kim2014convolutional}, the model prefers small filter size for text information. It is consistent with the experimental results in Fig. \ref{fig:filter_size}. When the filter size is set to (3,3), the F1-measure of the model is 0.92-0.93.



\section{Conclusions and Future Work}
The spread of fake news has raised concerns all over the world recently. These fake political news may have severe consequences. The identification of the fake news grows in importance. In this paper, we propose a unified model, i.e., TI-CNN, which can combine the text and image information with the corresponding explicit and latent features. The proposed model has strong expandability, which can easily absorb other features of news. Besides, the convolutional neural network makes the model to see the entire input at once, and it can be trained much faster than LSTM and many other RNN models. We do experiments on the dataset collected before the presidential election. The experimental results show that the TI-CNN can successfully identify the fake news based on the explicit features and the latent features learned from the convolutional neurons. 


The dataset in this paper focuses on the news about American presidential election. We will crawl more data about the France national elections to further investigate the differences between real and fake news in other languages. It's also a promising direction to identify the fake news with much social network information, such as the social network structures and the users' behaviors. In addition, the relevance between headline and news texts is a very interesting research topic, which is useful to identify the fake news. As the development of Generative Adversarial Networks (GAN) \cite{goodfellow2014generative,radford2015unsupervised}, the image can generate captions. It provides a novel way to evaluate the relevance between image and news text. 




\end{document}